\documentclass[sigconf]{acmart}
\usepackage{algorithm}
\usepackage{algorithmic}
\usepackage{booktabs} % For formal tables
\usepackage{multirow}
\AtBeginDocument{%
  \providecommand\BibTeX{{%
    \normalfont B\kern-0.5em{\scshape i\kern-0.25em b}\kern-0.8em\TeX}}}

\setcopyright{iw3c2w3}
\copyrightyear{2021}
\acmYear{2021}
\acmDOI{10.1145/XXXXXXX.XXXXXXX}

\acmConference[WWW '21 Companion]{Companion Proceedings of the Web Conference 2021}{April 19--23, 2021}{Ljubljana, Slovenia}
\acmBooktitle{Companion Proceedings of the Web Conference 2021 (WWW '21 Companion), April 19--23, 2021, Ljubljana, Slovenia}
\acmPrice{}
\acmDOI{10.1145/3442442.3452323}
\acmISBN{978-1-4503-8313-4/21/04}
\begin{document}

%%\title{The Name of the Title is Hope}
\title{Boosting Share Routing for Multi-task Learning}
\author{Xiaokai Chen}
\affiliation{\institution{Tencent PCG} \country{China}}
\email{xiaokaichen@tencent.com}

\author{Xiaoguang Gu}
\affiliation{\institution{Tencent PCG} \country{China}}
\email{ryanxggu@tencent.com}

\author{Libo Fu}
\affiliation{\institution{Tencent PCG} \country{China}}
\email{derekfu@tencent.com}
\renewcommand{\shortauthors}{Xiaokai and Xiaoguang, et al.}

\begin{abstract}
Multi-task learning (MTL) aims to make full use of the knowledge contained in multi-task supervision signals to improve the overall performance. How to make the knowledge of multiple tasks shared appropriately is an open problem for MTL. Most existing deep MTL models are based on parameter sharing. However, suitable sharing mechanism is hard to design as the relationship among tasks is complicated. In this paper, we propose a general framework called Multi-Task Neural Architecture Search (MTNAS) to efficiently find a suitable sharing route for a given MTL problem. MTNAS modularizes the sharing part into multiple layers of sub-networks. It allows sparse connection among these sub-networks and soft sharing based on gating is enabled for a certain route. Benefiting from such setting, each candidate architecture in our search space defines a dynamic sparse sharing route which is more flexible compared with full-sharing in previous approaches. We show that existing typical sharing approaches are sub-graphs in our search space. Extensive experiments on three real-world recommendation datasets demonstrate MTANS achieves consistent improvement compared with single-task models and typical multi-task methods while maintaining high computation efficiency. Furthermore, in-depth experiments demonstrates that MTNAS can learn suitable sparse route to mitigate negative transfer.
\end{abstract}




\maketitle

\section{Introduction} \label{sec1}
Multi-task learning (MTL) aims to make full use of the knowledge contained in multi-task labels to improve the overall performance\cite{caruana1997multitask}. It has been extensively used in many real-world applications, such as recommendation systems\cite{bansal2016ask,covington2016deep,zhao2019recommending}, e.g. predicting whether a user will click on a video and the engagement time the user will spend simultaneously. Compared with learning tasks separately MTL benefits a lot. It not only reduces maintenance cost of online systems but also could achieve better performance. MTL can be viewed as a form of inductive transfer, from this perspective, it generalizes better by introducing inductive biases contained in tasks\cite{ruder2017an}.  However, the inductive biases may conflict with each other, sharing information with a less related task in this case might actually hurt performance which is known as negative transfer\cite{ma2018modeling}. Therefore, a crucial problem in MTL is how to appropriately balance the sharing route of parameters for each task.

\begin{figure}[t]
\centering
\includegraphics[width=\linewidth]{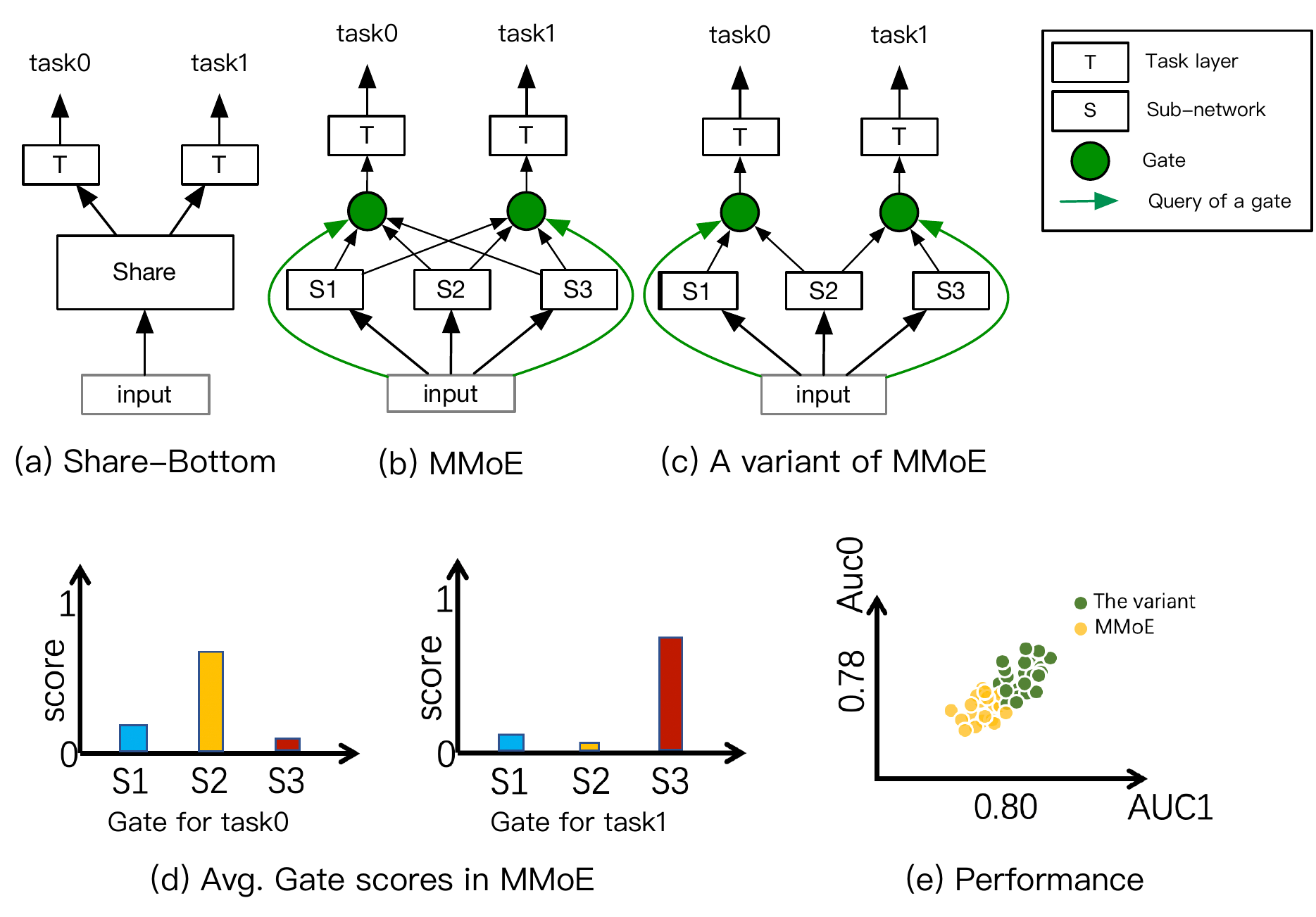}
\caption{(a)Share-Bottom. (b) MMoE. (c) A variant of MMoE. (d)Average gate scores in MMoE trained on BookCrossing\cite{ziegler2005improving}. (e) Comparison of the variant and MMoE on BookCrossing with two tasks, for each model, we have 20 runs with independent random initialization and the hyper-parameters are kept the same.}
\label{fig1}
\end{figure}

Most existing neural-based approaches achieves MTL via parameter sharing\cite{caruana1997multitask,ruder2017an,duong2015low,misra2016cross,yang2016deep}. As shown in Fig\ref{fig1}(a,b), there exist two typical architectures of parameter sharing, hard and soft sharing. A classical approach for hard sharing is called Share-Bottom\cite{caruana1993multitask}. It shares the bottom hidden layers across all task layers. Although it reduces the risk of overfitting, it suffers from significant degeneration in performance when the tasks are less related. This is due to its sharing strategy forces the input of all task-specific layers be the same, which harms model performance. Recent work MMoE\cite{ma2018modeling} is a representative soft sharing approach, it divides the shared bottom hidden layers into several parallel sub-networks and then aggregates the results of all sub-networks by gates. Compared with Share-Bottom, MMoE achieves more flexible parameter sharing.
%介绍两个代表性方法

Interestingly, in practice we find that a variant architecture consistently outperforms MMoE on a real-world recommendation dataset BookCrossing\cite{ziegler2005improving}, the comparison is shown in Fig\ref{fig1}(e). The variant is similar to MMoE, except for some connections between sub-networks and task layers are cut off, as shown in Fig\ref{fig1}(c). Note that the variant is a special case in MMoE when the corresponding weight scores given by gates are constantly zero. This motivates us to ask: why the variant is superior and why the gating mechanism is not enough to establish a better architecture at least similar to the variant? On the one hand, we find that MMoE is inherently a \textit{full-sharing} (in terms of topology) approach while the variant is a \textit{sparse} version. We argue that the full-sharing style of MMoE forces all experts to contribute to all tasks, which limits the flexibility of the sharing route. On the other hand, gating, as an idea to achieve dynamic feature aggregation, can hardly help learning a sparse connection although theoretically possible, as shown in Fig\ref{fig1}(d) some gate scores are very low, but they don't equal zero. 

The above analysis implies that there exist various sparse sharing routes which are beyond the capability of gates to establish. Therefore, we are concerned with two questions:(i) compared with full-sharing, is sparse connection generally helpful for MTL (e.g., mitigates negative transfer)? (ii) if so, how to find the best sparse sharing route automatically? 

%A solution is to manually tune the full-sharing architecture, e.g. cutting off some connections, however this requires prior knowledge of relationship among tasks, which is still an open problem\cite{}. Furthermore, the choice of query for a gate that used to achieve soft sharing is simply set as the root input features in previous work, we argue that the choice of queries is also important since features in different layers has different level of abstractions, while it has not been fully discussed yet. Therefore, designing a good architecture manually is challenging considering the potential sharing routes are numerous, thus how to automatically design an appropriate custom sharing route for a given MTL problem is worth studying.

\textbf{Present work.} To answer these two questions, we propose a general framework, called MTNAS(Multi-Task Neural Architecture Search) to efficiently find a suitable sharing route for a target MTL problem. Both sparse and full-sharing routes are considered in our search space. MTNAS modularizes the sharing part into multiple layers of sub-networks. In contrast to full-sharing in existing typical approaches, sparse connections among sub-networks are allowed, while soft sharing based on gating is enabled for a certain route. Besides, in order to make full use of soft sharing, the choice of query for a gate is also considered in the search space. Benefiting from this setting, each candidate architecture in our search space defines a certain dynamic sharing route, which is more general compared with previous work. Following ProxylessNAS\cite{cai2019proxylessnas}, a practical NAS approach, we directly learn the architecture parameters and weights for large-scale target MTL data, while the computational cost is in the same level of training a regular model. By comparing the learned sharing route with existing full-sharing approaches, the questions we concerned can be answered.

Search space is important, however, in MTL context, there exist no off-the-shelf operations (e.g., convolution, pooling) that widely used in visual domain\cite{zoph2016neural}. In order to balance sharing route, we first design a search space that is composed of multiple search blocks, each block enumerates all possible combinations of features from sub-networks in previous layer as well as all possible choices of the query for a gate. Based on ProxylessNAS, each candidate architecture is sampled from an over-parameterized network with multinomial distributions, then REINFORCE\cite{williams1992simple} is applied to optimize architecture parameters since it generalizes to non-differentiable evaluation metrics, while weights are tuned via standard gradient back propagation. At inference time, the sharing route and its weights are derived by picking up the local route with the maximum probability in each block. Extensive experiments on three real-world recommendation datasets demonstrate MTNAS achieves consistent improvement compared with single-task models and typical multi-task methods while maintaining high computation efficiency. Furthermore, in-depth experiments show that MTNAS is able to find a sparse route that can effectively alleviate negative transfer when tasks are less related.

\section{Related work}
\textbf{Parameter sharing in Multi-task learning.} There are a number of deep models that improve MTL via parameter sharing. Share-Bottom\cite{caruana1993multitask} is a classical approach which shares bottom hidden layers across all task layers, although it reduces the risk of overfitting, it suffers from negative transfer when tasks are less related, as it forces all tasks to share the same set of parameters in bottom layers. Recent work focus on designing flexible parameter sharing methods to mitigate negative transfer. L.Duong et al. \cite{duong2015low} adds L2 constraint between two single-sized model parameters, however it requires prior knowledge of task relatedness which is still an open problem\cite{baxter2000model,ben2003exploiting,ben2002theoretical}. I.Misra et al. \cite{misra2016cross} proposed a cross-stitch module that learns to fuse the features extracted from different single-task models. Compared with Share-Bottom, these models improve performance with more task-specific parameters. For efficiently serving, J.Ma et al. proposed MMoE\cite{ma2018modeling}, which splits the shared bottom layers into sub-networks and utilizes gates to aggregate all features of sub-networks for each task. However, the shared parts are all fully connected (in terms of topology) to task layers in these previous works. Our approach allows sparse connections among shared sub-networks, that is, parameters of different tasks are partially overlapped, thus our approach can find more flexible sharing routes while maintaining high computation efficiency.

\textbf{Neural architecture search(NAS).} NAS has become an increasing popular method to design architectures automatically. Reinforcement learning based methods\cite{zoph2016neural,baker2016designing} are usually computational intensive, an alternative to RL-based methods is the evolutionary approach\cite{liu2017hierarchical}, which optimizes the neural architecture by evolutionary algorithms. Some recent work\cite{brock2017smash,liu2018darts,cai2019proxylessnas} are proposed to accelerate NAS by one-shot setting, where the network is sampled by a hyper graph, and weight sharing is adopt to accelerate the search process. For instance, DARTS\cite{liu2018darts} optimizes the weights with a continuous relaxation, so that the architecture parameters can be updated via standard gradient descend. ProxylessNAS\cite{cai2019proxylessnas} further addresses the issue of large GPU memory consumption with path binarization. However, these previous work attempt to find a good architecture for a given task, the focus of this work is to find a suitable sharing route for an MTL problem, and consequently our search space is relatively small. Our approach is closely related to the SNR\cite{ma2019snr}, which learns connections among sub-networks. Nevertheless, the sharing route found by SNR is static, and it requires L0 regularization to achieve sparsity. In contrast, each specific sharing route in our search space is dynamic. What's more, sparsity is naturally considered in our search space, hence there is no need to tune hyper-parameters to control the sparsity in our MTNAS.

\begin{figure*}[!htbp]
\centering
\includegraphics[width=0.9\linewidth]{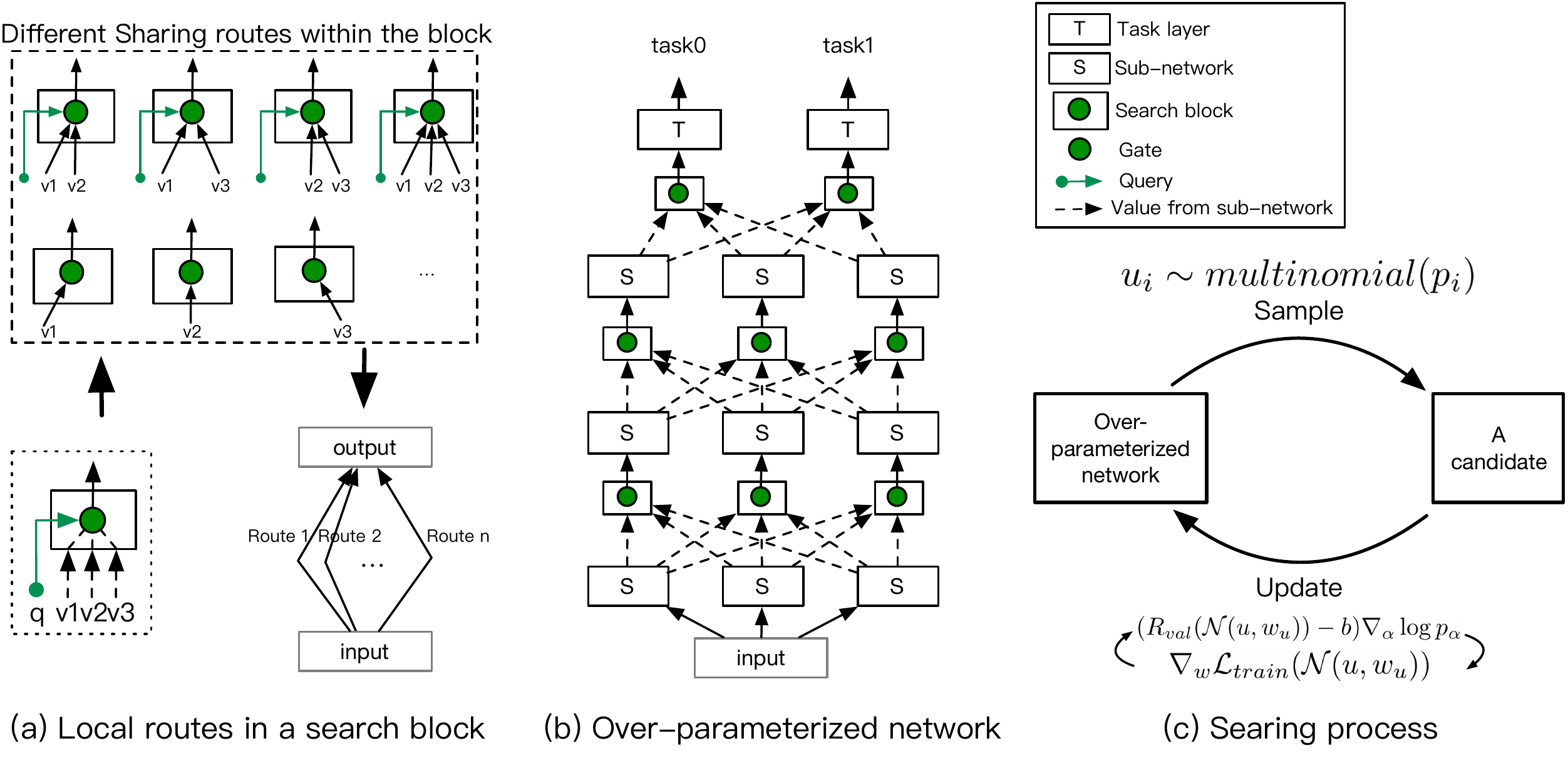}
\caption{The illustration of our approach MTNAS. (a) A block defines a sub-space, where different local sharing routes exist, intuitively each route represents a certain combination of inputs (totally 3 values in this example) that come from previous sub-networks. The choice of query for a gate in each route is also contribute to the sub-space. (b) The over-parameterized network defines our search space; it is composed of multiple search blocks (3 layers in this example). The query of a block is omitted for convenience. (c) An illustration of searching process. Each candidate sharing route is sampled from multinomial distributions determined by architecture parameters. The architecture parameters $\alpha$ and its weights $w_u$ are optimized alternately.}
\label{fig2}
\end{figure*}

\section{Proposed method: MTNAS}
In this section, we introduce MTNAS in details. The key idea is to allow sparse connections among shared sub-networks, that is, parameters can be partially shared across different tasks, while soft sharing is enabled for a specific route. By utilizing NAS to search a promising sharing route for an MTL problem, we can confirm whether sparsity in sharing route is generally beneficial for MTL. We first describe the search space (Section\ref{sec31}), where each candidate architecture defines a certain sharing route. Then we describe how the architecture and its weights can be learned via gradient-based method (Section\ref{sec32}). Finally, we describe the searching process and how to derive the learned route after searching (Section\ref{sec33}).
\vspace{-1em}
\subsection{Architecture search space: describing sparse and full-sharing routes in a unified form}\label{sec31}
The design philosophy behind our search space is to include both sparse and full-sharing routes so that the search space is general enough to solve MTL problems. MTNAS modularizes the sharing part into multiple layers of sub-networks. Without loss of generality, for a target MTL problem with $T$ tasks, we assume there exist $L$ sub-networks layers in the space, with each layer has $H$ sub-networks.

Following the typical block-wise NAS paradigm, our whole search space is composed of multiple search blocks. Each block defines a sub-space that contains different local sharing routes (e.g., connections among sub-networks) with each route defines a sub-path. Unlike most search blocks designed in visual domain\cite{liu2018darts,cai2019proxylessnas}, where each sub-path has its own weights, a sub-path in our search space contains hardly any weights except for a lightweight gate used to achieve dynamic aggregation, as shown in Fig\ref{fig2}(a). The intuition behind this setting is that our search space is generated by enumerating all local sharing routes and the sub-networks are shared during searching (Fig\ref{fig2}(b)). Note that we allow all blocks to be searched simultaneously. Next, we introduce the configuration of a block.

\textbf{The configuration of a block.} As shown in Fig\ref{fig2}(a), a block defines a sub-space, where different local sharing routes are contained, intuitively each route represents a certain combination of inputs that come from previous sub-networks. Besides, in order to achieve soft sharing for a certain route, we assign a gate to each route in the block. What's more, considering a query comes from different layers has different abstract levels, we set the source of a query can also be searched, so the choice of queries are also contribute to the sub-space. Formally, the sub-space can be represented as the Cartesian product $\mathcal{R}=\mathcal{V} \times \mathcal{Q}$, where $\mathcal{V}$ represents all possible combinations of values and $\mathcal{Q}$ represents all choices of a query. To be specific, $\mathcal{V}$ is a set that contains all possible combinations of outputs given by $H$ sub-networks in a previous layer $l$. The cardinality of set $\mathcal{V}$ is $|\mathcal{V}|=C_1^H+C_2^H+...+C_H^H=2^H-1$, where $C_i^H$ indicates the number of i-combinations from a set of $H$ values. A query $q\in \mathcal{Q}$ can come from the output of any sub-network in previous layer. There exist $|\mathcal{R}|$ different sharing routes within a block. For a certain route (e.g., $k^{th}$ route, $k=1,2,...,|\mathcal{R}|$), the inputs of a block are $\hat{V}$ and $\hat{q}$, where $\hat{V} \in \mathbb{R}^{s*d_v}$ indicates $s$ values with each has dimension $d_v$, and $\hat{q}\in \mathbb{R}^{d_q}$ indicates the query with dimension $d_q$, the output $y_k$ is the weighted sum of values: 

\begin{equation}
\begin{aligned}
y_k=g^k({\hat{q}}, \hat{V})=\sum_{i=1}^s m_i v_i \quad  \text{with} \\
 m_i=\frac{exp(e_{i})}{\sum_{i=1}^s exp(e_{i})}, e=w^k \hat{q} \label{eq1}
\end{aligned}
\end{equation}
where $g^k$ indicates the gate for k-th route, $m_i$ is the gate score for i-th value $v_i$, $w^k$ is the parameter in the gate to be learned.

\textbf{The over-parameterized network.} As shown in Fig\ref{fig2}(b), a block is located either between two consequence sub-network layers or between the last sub-network layer and task-layers. Consequently, the total number of blocks is $B=(L-1)*H+T$. Based on the above setting, the whole search space $\mathcal{A}$ can be represented as the Cartesian product of sub-spaces in blocks, $\mathcal{A}=\prod_{i=1}^{B} \mathcal{R}_i$, which is an over-parameterized network. This over-parameterized network is able to describe various sharing routes, since existing typical parameter sharing approaches (e.g., Share-Bottom and MMoE) are sub-graphs in the space.

\subsection{Search the sharing route with multinomial distribution learning}\label{sec32}
With the search space defined above, in this section, we introduce how to find a promising sharing route for a given MTL problem from the perspective of NAS.

Following ProxylessNAS\cite{cai2019proxylessnas}, we start from an over-parameterized network (supernet) from which a candidate sharing route can be sampled. The goal is to find a good candidate route in the supernet to perform MTL. Note that our search space is composed of $B$ blocks with each has $|\mathcal{R}_i|$ local routes, therefore, by selecting a local route in each block, a candidate sharing route can be established. Specifically, a candidate sharing route is denoted as $\mathcal{N}(u, w_u)$, where $u\in \mathbb{R}^B$ indicates the selecting actions for $B$ blocks and $w_u$ indicates its weights. 

%Firstly, we give a brief illustration of NAS. We denote $\mathcal{N}(graph_\alpha, w)$ as a network, where $graph_\alpha \in \mathcal{A}$ indicates the architecture determined by architecture parameter $\alpha$ and $w$ indicates the weights. The goal is to find an architecture $graph_{{\alpha}^*}$ that minimizes the validation loss $\mathcal{L}_{val}(\mathcal{N}(graph_{{\alpha}^*}, w^*))$, where $w^*$ is obtained by minimizing training loss. Thus NAS can be formalized as:
%\begin{equation}
%\mathop{\min}_{\alpha} \mathcal{L}_{val}(\mathcal{N}(graph_\alpha, w^*)), s.t., w ^*= \mathop{\arg \min}_{w} \mathcal{L}_{train}(\mathcal{N}(graph_{\alpha^*} , w)) \label{eq4}
%\end{equation}
%Eq\ref{eq4} implies a bi-level optimization problem, following \cite{liu2018darts,cai2019proxylessnas} we optimize it alternately.

\textbf{Architecture parameter optimization.} We set each action $u_i$ is sampled from a multinomial distribution determined by architecture parameters $\alpha_i \in \mathbb{R}^{|\mathcal{R}_i|}$:
\begin{equation}
u_i \sim multinomial({p_i}) \label{eq5}
\end{equation}
\begin{equation}
{p_i}=softmax({\alpha_i}) \label{eq6}
\end{equation}
Therefore a sharing route can be established by sampling from $B$ multinomial distributions. In order to optimize on non-differentiable evaluation metrics, we utilize REINFORCE\cite{williams1992simple} to optimize architecture parameters by maximizing the expected reward got from sampled architectures. The intuition behind the optimization is that the candidate sharing route with good performance should have a higher probability to be sampled. Therefore, the object function for optimizing architecture parameters are formalized as:
\begin{equation}
J({\alpha})=E_{{u}\sim {p(\alpha)}}R_{val}(\mathcal{N}(u, w_u)) \label{eq7}
\end{equation}

where ${p(\alpha)}$ indicates the multinomial distributions determined by architecture parameters ${\alpha}$, $R_{val}$ indicates the reward(e.g., AUC, accuracy, -loss) of the sampled architecture on validation set. According to REINFORCE\cite{williams1992simple}, the gradient of architecture parameters is obtained as follows:
\begin{equation}
\nabla_{{\alpha}} J({\alpha})=(R_{val}(\mathcal{N}(u, w_u))-b)\nabla_{{\alpha}} \log p_{{\alpha}} \label{eq8}
\end{equation}
where b is the baseline to reduce the variance of reward, here we simply compute the moving average of rewards experienced so far as baseline.

\subsection{Details in training and testing}\label{sec33}
In this section, we describe the training of architecture parameters and its weights (Fig\ref{fig2}(c)), and introduce the deriving of learned route after searching (Algorithm\ref{alg:A}). 

An instinctive understanding of Algorithm\ref{alg:A} is that at each iteration, a sharing route is sampled from multinomial distributions that determined by architecture parameters, then the architecture parameters as well as the weights of this sampled route are optimized in turn, and as this process iterates, the sampling probabilities of routes with good performance gradually increase. After searching, we pick up the local sharing route with the maximum probability in each block to derive the final architecture.

%\subsection{Self-boosting via adaptive baseline}
%The major limitation of REINFORCE is that the expected reward got from samples typically exhibit high variance. "Baseline" is usually used to normalize the reward and reduce the variance, as stated in Eq\ref{eq8}. Moving average of rewards experienced so far is a straightforward method to compute baseline. Here, following SCST\cite{rennie2017self} we propose an adaptive baseline to boost the searching process.
%
%Specifically, we set the baseline as the reward got from an architecture $\mathcal{N}(graph_{\hat{action}}, w)$ sampled with greedy strategy, where $\hat{action}_i=\mathop{\arg \max}_j p_i$. This method adaptively normalize the reward with currently best architecture, therefore only sampled architectures that outperform the greedy architecture are given positive weight, and inferior samples are suppressed. Then Eq\ref{eq8} can be updated as:
%\begin{equation}
%\nabla_{{\alpha}} J({\alpha})=(R_{val}(\mathcal{N}(graph_{{action}}, w))-R_{val}(\mathcal{N}(graph_{\hat{action}}, w)))\nabla_{{\alpha}} \log p_{{\alpha}} \label{eq9}
%\end{equation}
%
%We show the overall algorithm in Alg\ref{alg:A}. We sample architectures and then alternatively optimize the weights and the architecture parameters. After searching, we pick up the sharing route with the maximum probability in each block to derive the final architecture.

\begin{algorithm}
\caption{Architecture searching and deriving}
\label{alg:A}
\begin{algorithmic}
\REQUIRE train set and valid set, supernet with $B$ blocks.
\ENSURE Optimized architecture parameters $\alpha$ and weights $w$.
\STATE Initialize $\alpha$ and $w$.
\WHILE {not converged}
\FOR{each $\text{block}_i$ in supernet}
\STATE Calculate the sample probability for different local sharing routes via Eq\ref{eq6};
\STATE Sample a local sharing route $u_i$ via Eq\ref{eq5};
\ENDFOR
\STATE Get a candidate sharing route $\mathcal{N}(u, w_u), \ \text{with} \ u=\{u_i\}_{i=1}^B$;
\STATE Update weights $w$ by descending gradient computed via $\nabla_w \mathcal{L}_{train}(\mathcal{N}(u, w_u))$;
\STATE Update architecture parameters $\alpha$ by ascending gradient computed via Eq\ref{eq8};
\ENDWHILE 
\RETURN Derive the final architecture based on $\alpha$ and $w$.
\end{algorithmic}
\end{algorithm}

\section{Experiments}
We design our experiments with the goals of (i) verifying the improvement of MTNAS over typical sharing approaches and (ii) answering whether sparsity in sharing route is generally helpful to mitigate negative transfer. We first conduct experiments on three real-world recommendation datasets, considering there naturally exist multiple tasks in recommendation systems. We then conduct experiments on synthetic data to further confirm the effectiveness of MTNAS.
\vspace{-1em}
 
 \begin{table*}[!htbp]
%%\vspace{-1.5em}%%%%%%%%%%%%%%%%%%%%缩减竖直距离%%%%%%%%%%%%%%%%%%%%%%
\centering
\caption{Statistics of the datasets.}
\renewcommand\tabcolsep{1.8pt} % 调整表格列间的宽度
\begin{tabular}{ccccccc}
\hline
Dataset& \#Train& \#Valid &\#Test& Fields& Features(Sparse)&Task number\\
\hline
Bookcrossing& 800,000& 200,000&149,780& 15&1,000,000&2\\
Tiktok& 15,697,872& 1,962,234&1,962,234&9&4,000,000&2\\
Goodreads& 1,822,664& 455,666&456,020&14&500,000&3\\
\hline
\end{tabular}
\label{tb1}
\end{table*}
%%\vspace{-1em}

\begin{table*}[]
\centering
\caption{Results on BookCrossing and Tiktok (higher AUC is better). The performance degradation compared with single-task model is in gray.}
\begin{tabular}{c|ccc|ccc}
\hline
\multirow{2}{*}{Method} & \multicolumn{3}{c|}{BookCrossing} & \multicolumn{3}{c}{Tiktok} \\ \cline{2-7} 
                              & AUC0      & AUC1     & MTL-Loss   & AUC0    & AUC1   & MTL Loss \\ \hline
\multirow{2}{*}{Single}                              & 0.7842    & -        & -          & 0.7485  & -      & -        \\ 
                        & -         & 0.7984   & -          & -       & 0.9428 & -        \\ \hline
Share-Bottom                  & 0.7834    & 0.8014   & 0.7322     &  0.7478  &  0.9415 & 0.6020   \\ 
MMoE                          & 0.7885    & 0.8022   & 0.7302     & 0.7488  & 0.9425 & 0.6006   \\ 
ML-MMoE                       & 0.7884    & 0.8051   & 0.7299     & 0.7487  &  0.9421 & 0.6011   \\ \hline
\textbf{AutoMTL(Ours)}                         & \textbf{0.7907}    & \textbf{0.8086}   & \textbf{0.7247}     & \textbf{0.7507}  & \textbf{0.9467} & \textbf{0.5930}   \\ \hline
\end{tabular}
\label{t2}
\end{table*}

\subsection{Experimental setup}\label{sec41}
\textbf{Datasets.} We evaluate our method on three public real-world recommendation datasets: Bookcrossing\cite{ziegler2005improving}, Goodreads\cite{wan2018item} and Tiktok\cite{tiktok}, datasets are randomly split into training, validation and testing sets, the statistics are listed in Table\ref{tb1}. We hash each feature into a hexadecimal number, and calculate the remainder to get the feature id. Bookcrossing contains 278,858 users providing 1,149,780 ratings about 271,379 books. We construct a MTL problem with two tasks: predicting whether a user rates a book (denoted task0) and whether a user is satisfied with a book(task1). Goodreads is collected from a book review website. Due to the complete dataset is extreame large, we use a medium-size subset by setting genre as poetry following the official advise. It contains 2,734,350 interactions collected from 377,799 users' public bookshelves and covers 36514 books. An MTL problem with three tasks is constructed by predicting whether a user reads a book(task0), whether a user is satisfied(task1) and whether a user leaves a review text(task2). Tiktok is a dataset provided by Bytedance in the 2019 Short Video Understanding Challenge. There are two tasks in the challenge, predicting whether a user finishes watching a video(task0) and whether a user likes a video(task1). 

\textbf{Evaluation metrics.} Since each single task is a binary classification problem, the cross-entropy loss is used for each task. Therefore, the total loss (denoted MTL-Loss) for MTL of each dataset is a sum of loss for each single task. We use AUC (area under ROC curve) scores as evaluation metric. It is noticeable that a slightly higher AUC at 0.001-level is regarded significant for CTR-prediction like tasks, which has also been pointed out in existing works\cite{cheng2016wide, guo2017deepfm,wang2017deep}. 
%We utilize Pearson correlation coefficient to estimate the relationship of tasks to help analysis.
%https://biendata.com/competition/icmechallenge2019/
%bookcrossing http://www2.informatik.uni-freiburg.de/~cziegler/BX/
%goodread https://sites.google.com/eng.ucsd.edu/ucsdbookgraph/home

\subsection{Implementation Details}
\textbf{Input feature} means the root shared feature. All the raw features including user-side and item-side are first discretized, then an embedding layer is used to map these one-hot features to distributed representations, then representations of different fields are concatenated to form the input feature. Embedding dimension is empirically set 10 for all datasets. 
%For the sake of simplicity and focusing the comparison of MTL performance we adopt plain DNN as backbone in all experiments. Note that our framework can be easily extended to other models such as wide\&deep, DeepFM and so on. The output of backbone is the input feature for different MTL methods.

\textbf{Methods.} We compare MTNAS with three widely used approaches: Share-Bottom\cite{caruana1993multitask}, MMoE\cite{ma2018modeling} and ML-MMoE. There are two fully-connected layers in Share-Bottom. As for MMoE, we set one fully-connected layer followed by four sub-networks, each of which being a one-layer fully-connected layer. ML-MMoE is an extension of MMoE, which has multiple sub-network layers. As for MTNAS, we set one fully-connected layer followed by two subsequent sub-network layers($L=2$) with each layer having four sub-networks($H=4$), each of which being a one-layer fully-connected layer. In MTNAS, the hidden size of all fully-connected layers is set 64 and $R_{val}$ is set as the sum of AUC scores of each task. To make the comparison fair, we set the number of parameters approximately the same as MTNAS for other methods. Task layer in these models are all a fully connected layer. ReLU\cite{nair2010rectified} activation function is used wherever needed. As for single-task model, its architecture is the same as Share-Bottom except for only one task layer in it.

\textbf{Hyper-parameters.} With regard to MTNAS, for the three datasets, adam\cite{kingma2014adam} optimizer with an initial learning rate of 0.01 is used to update architecture parameters, the mini-batch size is 1024. For the optimization of weights, another adam optimizer with an initial learning rate of 0.001 is adopted, and weight decay is set 0.0005, the mini-batch size is 512. As for the compared methods, we apply grid search for optimal training hype-parameters. In order to provide a fair comparison, the loss function is kept the same for all methods. The Adam optimizer is also adopted to update parameters for all these methods.

\begin{table}[t]
\renewcommand\tabcolsep{2.pt} % 调整表格列间的宽度
\caption{Results on GoodRead (higher AUC is better).}
\begin{tabular}{ccccc}
\hline
Method& AUC0& AUC1 &AUC2&MTL Loss\\
\hline
\multirow{3}{*}{Single}& 0.8104& -&-&-\\
& -& 0.7752&-&-\\
& -& -&0.8209&-\\
\hline
Share-Bottom& 0.8250&  0.7748&0.8415&1.0726\\
MMOE& 0.8255& 0.7761&0.8441&1.0716\\
ML-MMOE& 0.8244&  0.7743&0.8443&1.0720\\
\hline
\textbf{AutoMTL(ours)}& \textbf{0.8281}& \textbf{0.7771}&\textbf{0.8460}&\textbf{1.0656}\\
\hline
\end{tabular} 
\label{t3}
\end{table}

\begin{figure*}[t]
\centering
\includegraphics[width=0.78\linewidth]{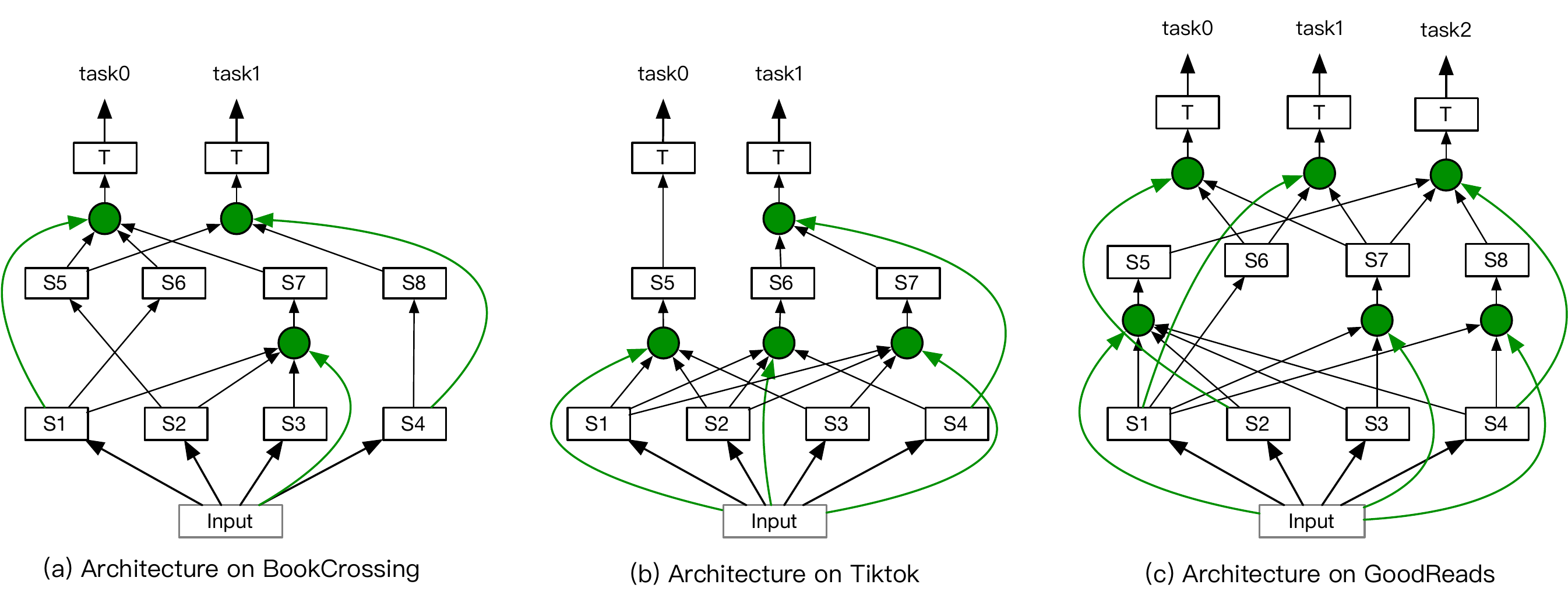}
\caption{The learned sharing routes on three datasets.}
\label{fig3}
\end{figure*}
%\vspace{1em}

%%\vspace{-1em}
\subsection{Main results} %Tiktok dataset(colored in gray) Especially
\textbf{Performance.} We compare our approach MTNAS with single task model as well as typical multi-task approaches, Share-Bottom, MMOE and ML-MMOE. The results on three datasets are listed in Table\ref{t2} and Table\ref{t3}. It can be observed that MTNAS consistently outperforms compared methods on three datasets, MTNAS achieves the highest AUC score on all tasks for three MTL problems, this demonstrates the effectiveness of MTNAS. We notice that negative transfer happens for compared multi-task methods (colored in gray), where the performance is worse than single-task model. For the ease of understanding, we compute Pearson correlation coefficient (PCC) to estimate the correlation of tasks, results are illustrated in Table\ref{t4}. It is observed that the PCC of two tasks in Tiktok is only 0.030, which is very close to zero, means that linear correlation of these two tasks is quite weak. Therefore, negative transfer is leaning to happen if the sharing route is not appropriate. However, MTNAS escapes from negative transfer, this indicates that it's able to find a suitable sharing route to make full use of knowledge of tasks. To further understand MTNAS, we discuss in Section\ref{s44}.

\textbf{The learned sharing route of MTNAS} is illustrated in Fig\ref{fig3}. In general, it can be observed that all the searched sharing routes are sparse (compared to full-sharing). As is shown in Fig\ref{fig3}(a), task-0 and task-1 share path \textit{input-s2-s5}, while other parts are private for each task. We suppose that this architecture provides a trade-off between the shared and independent parts, making each task take what it needs adaptively. As for Goodreads (Fig\ref{fig3}(c)), it's worth noting that task-0 and task-1 share the same combination of sub-networks: \textit{s6\&s7}. Interestingly, we find that the PCC of task-0 and task-1 is highest, 0.493, among three tasks (Table\ref{t4}). This phenomenon further suggests that our approach MTNAS can find interpretable sharing route for a given MTL problem.

\begin{table}[t]
\caption{The Pearson correlation (PCC) of tasks. t0 indicates task-0(defined in Sec\ref{sec41}).}
\begin{tabular}{|l|l|l|l|}
\hline
\multicolumn{1}{|c|}{{Dataset}} & t0\&t1 & t1\&t2 & t0\&t2 \\ \hline
BookCrossing                                                 & 0.348  & -      & -      \\ \hline
Tiktok                                                 & 0.030  & -      & -      \\ \hline
GoodReads                                                 & 0.493  & 0.124  & 0.245  \\ \hline
\end{tabular}
\label{t4}
\end{table}

\begin{table}[t]
\centering
\caption{Comparison of variants. We compare testing loss on the three datasets.}
\begin{tabular}{cccc}
\hline
Variants & BookCrossing & Tiktok & GoodReads \\ \hline
$\text{Baseline}$  &0.7286& 0.5961&1.0690       \\
$\text{MTNAS}_{\text{w/o QS}}$ &0.7260&0.5941&{1.0669}       \\ 
MTNAS  & \textbf{0.7247}& \textbf{0.5930}&\textbf{1.0656}\\\hline
\end{tabular}
\label{t5}
\end{table}

%so .  tease apart the query selection in search space and all queries are set as the root input feature. 
\textbf{Effectiveness of Components.} We compare MTNAS with two variants. \textit{Baseline} replaces gate with mean pooling, so the learned route is static. \textit{MTNAS(w/o QS)} teases apart the query selection and all queries are set as the root feature. As shown in Table\ref{t5}, our entire method outperforms all the variants. By comparing MTNAS with \textit{Baseline}, we confirm the effectiveness of dynamic aggregation. Compared to \textit{MTNAS(w/o QS)}, MTNAS achieves consistent improvements which demonstrates that it's able to find suitable queries and the root input feature is not always the best choice for query.

%\begin{figure}[t]
%\centering
%\subfigure[The average entropy of architecture parameters.]{
%\includegraphics[width=0.4\textwidth]{./imgs/img4a} 
%}
%\subfigure[The validation loss.]{
%\includegraphics[width=0.4\textwidth]{./imgs/img4b}
%}
%\caption{Influence of adaptive baseline.}
%\label{fig4}
%\end{figure}

% \usepackage{multirow}

%\caption{Ablation study comparing the performance of MTNAS with and without query selection. $\text{MTNAS}_{\text{w/}}$ is the complete model while the $\text{MTNAS}_{\text{w/o}}$ is the model without query selection.}
\section{Analysis}\label{s44}
%\textbf{Influence of sparse connection.}
%Sub-networks between two adjacent layers are fully connected in MMoE-like methods. Allowing sparse connection is an important feature in our search space. To justify its contribution, we replace it with fully connection and keep other settings as they are. As shown in Table*,  we observe that the performance is decrease on all three datasets if sparse connection in search space is removed. This indicates sparse connection is crucial to trade-off sharing route for MTL.

\subsection{Experiments on synthetic data.} 
To further make it clear when MTNAS works and how it behaves, we conduct experiments on synthetic data, where task relatedness $\rho$ can be controlled. We push it to an extreme case where $\rho=0$, in this case tasks are unrelated, so we can observe how MTNAS behaves. We generate two regression tasks; the synthetic data generation is illustrated below.

\textbf{Synthetic data generation} Following \cite{ma2018modeling}, the synthetic data is generated through the following four steps.

\begin{enumerate}
\item Given the input feature dimension $d$, we generate two orthogonal unit vectors $u_1,u_2\in \mathbb{R}^d$. To be specific,
\begin{equation}
u_1^Tu_2=0,||u_1||_2=|u_2||_2=1
\end{equation}
\item Given a correlation score $-1\le \rho \le 1$, we generate two weight vectors $w_1,w2$,
\begin{equation}
w_1=u_1,\ w_2=\rho u_1+ \sqrt{1-\rho^2} u_2
\end{equation}
with this setting, the cosine similarity of the these two weights is exactly $\rho$, that is, $cos(w_1,w_2)=\rho$.
\item Generate inputs. Randomly sample an input data point $x\in \mathbb{R}^d$, with each entry from a normal distribution $\mathcal{N}(0,1)$.
\item Generate two labels $y_1, y_2$ for two regression tasks as follows,
 \begin{equation}
y_1=w_1^T x+\sum_{i=1}^m sin(\alpha_i w_1^T x+\beta_i) + \epsilon_1
\end{equation}
 \begin{equation}
y_2=w_2^T x+\sum_{i=1}^m sin(\alpha_i w_2^T x+\beta_i) + \epsilon_2
\end{equation}
the sum of sine functions here make the mapping from input to label non-linear. $\epsilon$ is the noise from $\mathcal{N}(0,0.01)$. 
\end{enumerate}

Note that, the correlation parameter $\rho$ represents the cosine similarity of two task labels. Due to the non-linear data generation procedure, it's not straightforward to control the Pearson correlation coefficient (PCC). So we indirectly control PCC through $\rho$. We set $\rho=0$ and repeat these four steps 10,000 times to generate a dataset with two unrelated tasks. In our experiment, the PCC of two tasks is -0.0162.
\begin{figure}[htbp]
\centering
\includegraphics[width=5.5cm]{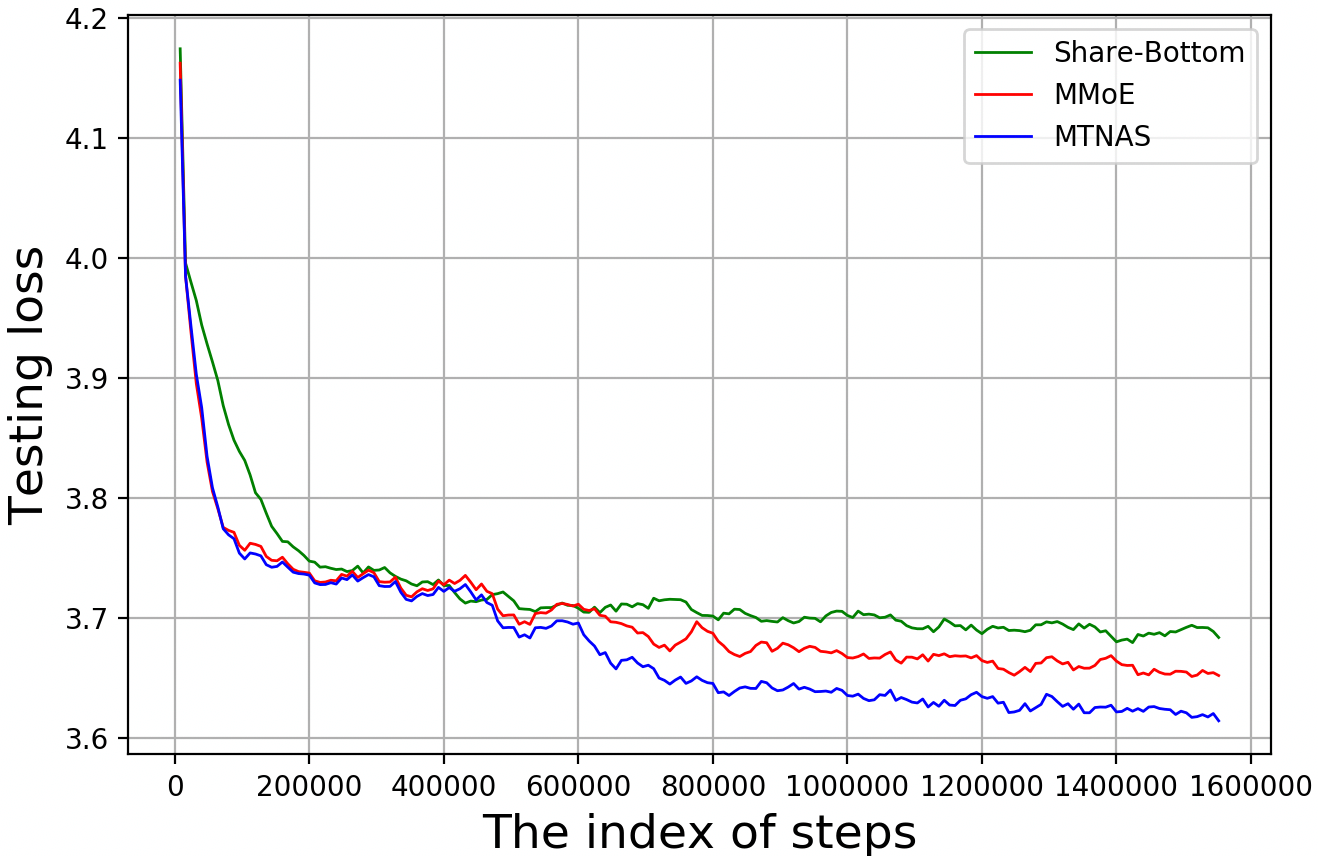}
\caption{Comparison of Share-Bottom, MMoE and MTNAS on synthetic data with two unrelated tasks. The plot show total loss over steps.}
\label{fig5}
\end{figure}

\begin{figure}
\centering
\centering
\includegraphics[width=2.5cm]{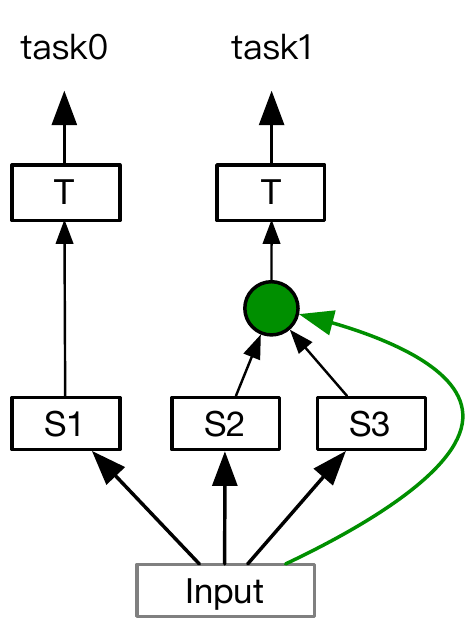}
\caption{The learned sharing route with L=1,H=3.}
\label{supfig1}
\end{figure}

\begin{figure}
\centering
\includegraphics[width=6cm]{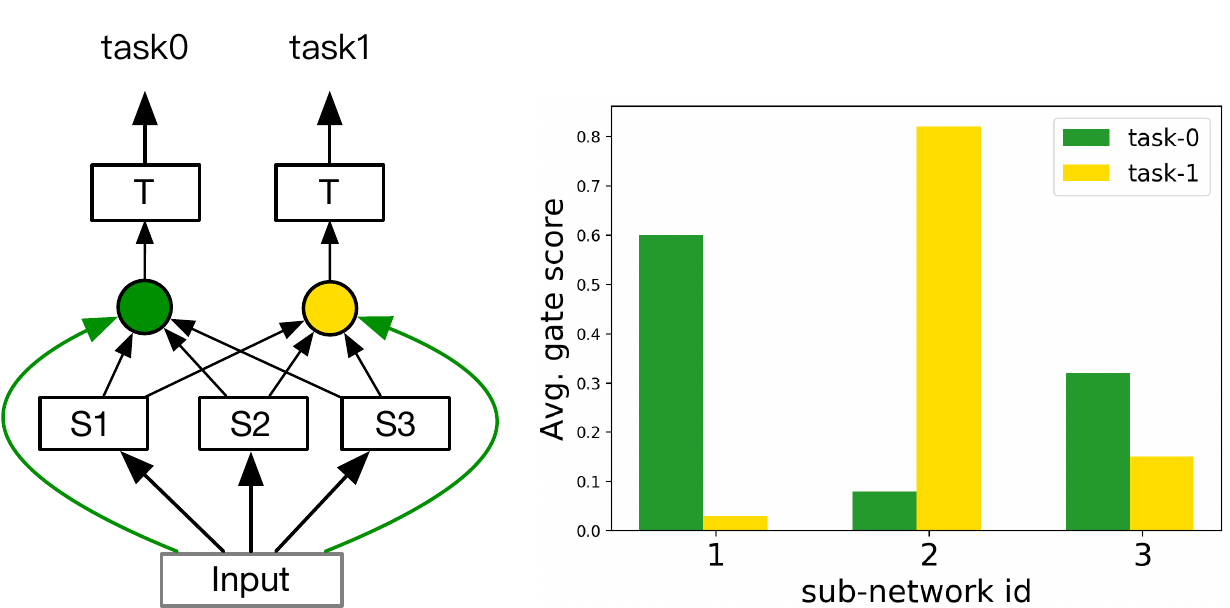}
           \captionsetup{justification=centering}
\caption{Left, the architecture of MMoE. Right, the distribution of average gate scores.}
\label{supfig2}
\end{figure}
\textbf{Results.} We compare Share-Bottom, MMoE with architecture searched by MTNAS on the synthetic data. To make the comparison fair, we set the number of parameters for each model approximately the same. The results are shown in Fig\ref{fig5}. It can be observed that MTNAS achieves the lowest loss. Note that the main difference between MMoE and MTNAS is that the sharing route in MMoE is fully connected (in terms of topology), while MTNAS's is sparse. Furthermore, a key feature of the learned sharing route is that \textit{no} overlap exists between two sharing route of each task, on the one hand this indicates that MTNAS behaves reasonable since no knowledge could be shared with two unrelated tasks, on the other hand we believe this demonstrates that sparsity is important to alleviate negative transfer when tasks are less related, because it allows parameters of each task to be partially shared, while although the average gate scores in MMoE for some shared parts are very low, it is still not equal to constantly zero.

%query的来源也涵盖在了我们的搜索空间中，为了验证query来源选择的有效性，我们比较了query固定取自input，与可以取自前层任意子网络的输出这两种情况，如**所示。把query纳入搜索空间的效果更好。这表明，来自不同层级的query的含有不能层面的指导信号，选择合适的query会进一步提升效果。

%\textbf{Influence of adaptive baseline.}
%To evaluate the effectiveness of adaptive baseline, we compare it with moving average baseline on Goodreads. We utilize average entropy of multinomial distributions to evaluate the convergence of architecture parameters $\alpha$ and the validation loss of best architectures is used to evaluate the performance. Results are shown in Fig *, it can be observed that (1) the average entropy of MTNAS $\alpha$ is lower than the variant, which means adaptive baseline can accelerate searching process, (2) MTNAS achieves lower loss than the variant, this confirms adaptive baseline provides a high quality normalization of rewards and can boost the search performance.
%为了验证自适应baseline的效果，我们将其与滑动平均进行对比。如*所示，自适应方法使得网络结构参数收敛速度加快。且最终结果也更好。可以鼓励算法搜索到比当前

\subsection{Parameter sensitivity.} 
\begin{figure}[htbp]
\centering
\includegraphics[width=5.8cm]{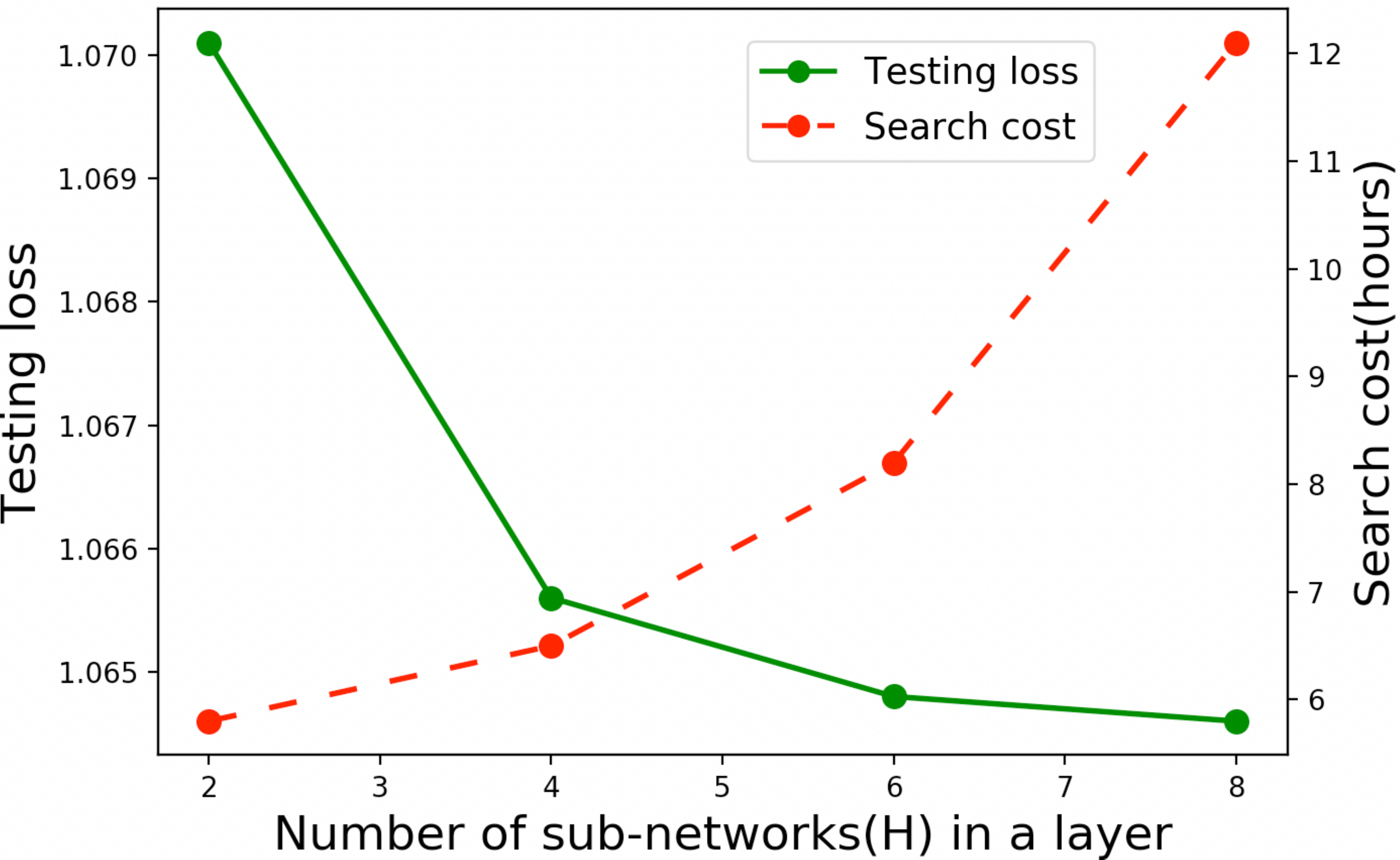}
           \captionsetup{justification=centering}
\caption{Performance on GoodReads w.r.t the number of sub-networks in a layer.}
\label{fig6}
\end{figure}

We study the sensitivity to $L$ and $H$(Sec\ref{sec31}). We find that setting $L=2$ provides good performance on the three datasets, while increasing $L$ beyond 2 gives marginal returns in performance. We also find diminishing returns and increasing search cost for larger $H$(Fig\ref{fig6}).

%\vspace{-1em}
\section{Conclusion}
We present MTNAS to efficiently find a suitable sparse sharing route to perform multi-task learning. MTNAS consistently outperforms single-task model and typical multi-task approaches on three real-world datasets, which indicates that it can establish a promising sharing route to make full use of the knowledge of tasks. Visualization of learned routes confirm that sparsity is beneficial for MTL as it allows parameters to be partially shared across tasks. Experiments on synthetic data further demonstrates that, by allowing sparse connection among shared sub-networks, MTNAS is able to find a sparse route that can effectively alleviate negative transfer when tasks are less related.

\bibliographystyle{ACM-Reference-Format}
\clearpage
\bibliography{sample-base}

\end{document}